\documentclass[letterpaper, 10 pt, journal, twoside]{IEEEtran}
\usepackage{amsmath,amsfonts}
\usepackage{algorithmic}
\usepackage{algorithm}
\usepackage{array}
\usepackage[caption=false,font=normalsize,labelfont=sf,textfont=sf]{subfig}
\usepackage{textcomp}
\usepackage{stfloats}
\usepackage{url}
\usepackage{verbatim}
\usepackage{graphicx}
\usepackage{cite}

\usepackage{mathtools}
\usepackage{amsmath}
\usepackage{caption}

\graphicspath{{figs/}}

\let\oldtwocolumn\twocolumn
\renewcommand\twocolumn[1][]{%
	\oldtwocolumn[{#1}{
		\begin{center}
			\includegraphics[width=\textwidth]{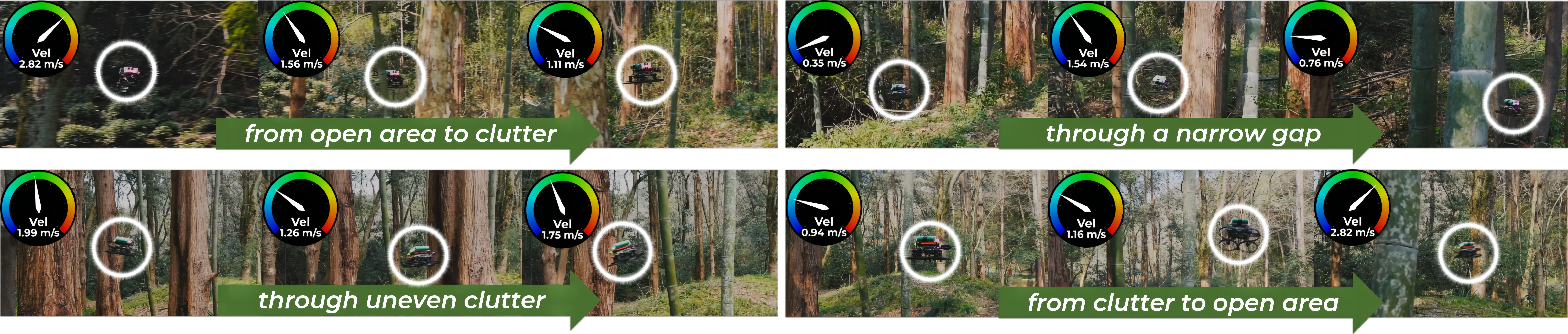}
				\captionsetup{font=footnotesize}  
			\captionof{figure}{\textbf{Flight with speed adaptation in complex natural clutter}. (Top left) The vehicle aggressively flies until it is near clutter, and further decelerates when observing a hidden bamboo behind a tree. (Top right) The vehicle flies cautiously when crossing a narrow gap between two tree trunks. (Bottom left) The vehicle flies smoothly between obstacles. (Bottom right) The vehicle flies aggressively after observing the open space in front of it.}
			\label{fig:head}
		\end{center}
	}]
}

\begin{document}

\title{Learning Speed Adaptation for Flight in Clutter}

\author{Guangyu Zhao$^{*}$, Tianyue Wu$^{*}$, Yeke Chen and Fei Gao 
\thanks{Manuscript received: March, 9, 2024; May, 25, 2024; June, 19, 2024.}
\thanks{This paper was recommended for publication by Editor Jens Kober upon evaluation of the Associate Editor and Reviewers' comments. This work was supported by the National Natural Science Foundation of China under grant no. 62322314. (\emph{Corresponding Author: Fei Gao})} 
\thanks{$^{*}$Guangyu Zhao and Tianyue Wu contribute this work equally.}%
\thanks{All authors are with the Institute of Cyber-Systems and Control, College of Control Science and Engineering, Zhejiang University, Hangzhou 310027, China, and also with the Huzhou Institute, Zhejiang University, Huzhou 313000, China.}%
\thanks{E-mail:{\tt\small \{zhaoguangyu, tianyueh8erobot, chenyeke, fgaoaa\}@zju.edu.cn}}
\thanks{Digital Object Identifier (DOI): see top of this page.}
}

\markboth{IEEE ROBOTICS AND AUTOMATION LETTERS. PREPRINT VERSION.ACCEPTED JUNE, 2024}
{Learning Speed Adaptation for Flight in Clutter} 


\maketitle

\begin{abstract}
		Animals learn to adapt speed of their movements to their capabilities and the environment they observe. Mobile robots should also demonstrate this ability to trade-off aggressiveness and safety for efficiently accomplishing tasks. The aim of this work is to endow flight vehicles with the ability of speed adaptation in prior unknown and partially observable cluttered environments. We propose a hierarchical learning and planning framework where we utilize both well-established methods of model-based trajectory generation and trial-and-error that comprehensively learns a policy to dynamically configure the speed constraint. Technically, we use online reinforcement learning to obtain the deployable policy. The statistical results in simulation demonstrate the advantages of our method over the constant speed constraint baselines and an alternative method in terms of flight efficiency and safety. In particular, the policy behaves perception awareness, which distinguish it from alternative approaches. By deploying the policy to hardware, we verify that these advantages can be brought to the real world.
\end{abstract}

\section{Introduction}

\label{sec:intro}
\IEEEPARstart{A}{nimals} seldom move at their maximum speeds due to limited sensory, reaction, or motor capabilities. For instance, some animals, such as birds, enhance the resolution of spatial perception by slowing down while foraging \cite{yoon2018control}. Cheetahs almost never chase their prey at full speed due to the difficulties of sharp turns or footing maintenance \cite{wilson2013locomotion}. Such compromising behaviors are especially likely to occur in constrained environments \cite{wheatley2015fast}, where animals regulate their speed to ensure safety considering their limited capabilities, e.g., budgerigars fly at a low speed to ensure collision-free crossing of narrow gaps \cite{henningsson2021flying}. 

The same goes for mobile robots. The limited sensory update frequency, time-consuming decision-making, and imperfect motor control capabilities, to name a few, inherently limit the allowed  speed of their movements that satisfies safety regards. This efficiency-safety trade-off is always present, no matter how far a hardware or algorithmic system has evolved.  Therefore, like animals, robots should be able to adaptively regulate their speed of movements based on an integrated cognition relating self-awareness, e.g., of their own capability limitations, and other-awareness, e.g., of the external environments \cite{smith2003comparative}.

This paper focuses on flight vehicles, where some agile behaviors have been preliminarily achieved \cite{zhou2019robust,o2022neural,song2023reaching,kaufmann2023champion}. These advances are made possible by the considerable development of model-based trajectory planners \cite{zhou2019robust,wang2022geometrically} and controllers \cite{o2022neural,sun2022comparative} in the last decades, and the recent application of model-free learning techniques \cite{song2023reaching,kaufmann2023champion}. However, most of the existing planning and control schemes leave the task of determining the speed constraint for the user, which is \emph{conservatively set to constant} at deployment time \cite{zhou2019robust,zhou2020ego}. While some works utilize reinforcement learning (RL) to learn a policy that directly outputs low-level commands with implicit speed adaptations \cite{song2023reaching,kaufmann2023champion}, the success of these works occurs for the time being only in prior known environments. In contrast, this paper considers the problem of safe flight in unknown, partially observed, and cluttered environments.

An intuitive idea is to trial and error, learning a policy \emph{from scratch} to enable naturally embedded speed adaptation according to the inherent limitations of the system, as implemented by a concurrent work \cite{yu2024mavrl}. However, such an approach is expensive to learn a deployable policy, and at this stage can only be deployed for simpler scenarios than a state-of-the-art model-based trajectory planner can handle \cite{zhou2019robust,zhou2020ego}. Instead, in this paper, we take advantage of the insights gained over the past decades in classical trajectory generation and tracking frameworks, which are favored due to their formal safety guarantee and generalizability that is not available by policies learned from scratch.

In particular, we decompose the policy into a hierarchical one, where the outer-loop policy dynamically configures the \emph{speed constraint} and can be effectively learned with online RL. The inner-loop policy, which is conditioned on the output of the outer-loop policy, generates the trajectory for execution. Fortunately, modern model-based trajectory planners \cite{zhou2019robust,zhou2020ego,loquercio2021learning}, despite their requirement for a pre-determined constant speed constraint, have evolved considerably to serve as a near-optimum to the inner-loop policy in the scenarios they are specifically designed for. Similar hierarchical frameworks that combine experience-based learning and model-based optimization can be found in problems such as model predictive control (MPC)~\cite{sacks2022learning,song2022policy,romero2023weighted} and legged locomotion \cite{yang2023cajun,jenelten2024dtc}.

Our main contribution is a system that hierarchically combines the model-based trajectory planner and a learned outer-loop policy to enable aggressiveness-adaptive flight in clutter. This approach is benchmarked to outperform baselines with constant constraints and an alternative approach \cite{quan2021eva}. We also demonstrate the system in real-world scenarios, including a complex natural clutter shown in Fig. \ref{fig:head}, exhibiting aggressive but safe flight in the wild. One of the crucial technical designs to effectively learn the outer-loop policy is a two-stage reward scheme that is employed to overcome the challenge of stochasticity and sparsity posed by the early-termination penalty. \vspace{-0.3cm}

\section{Related Work}

\subsection{Adaptive Motion Planning}

\label{sec:2A}
The works that have the most similar motivation to us are \cite{quan2021eva,zhou2022automatic,wang2023speed}, which endeavor to alter flying speed with model-based motion planners. In \cite{quan2021eva,wang2023speed}, the authors impose the desired adaptive behavior by incorporating the velocity into trajectory planning via handcrafted cost functions. However, these cost functions are not expressive enough to comprehensively address the problem and inevitably lead to inflexible behavior. Methodologically, while we believe that powerful modeling-based approaches can be raised through more generalized and accurate formulations and specialized computational techniques, learning-based approaches, if feasible, can serve as a simpler alternative. Zhou et al. \cite{zhou2022automatic} employ an online learning method based on Beyasian optimization to decide hyperparameter for trajectory planners. However, the painful convergence speed of Beyasian optimization makes their method incapable of adapting to a rapidly updating observation. We believe that an offline training mode, where the policy is determined before deployment, is more appropriate for the goal. Richter et al. \cite{richter2018bayesian} formalize the problem of planning in an unknown environment and indicate the fundamental intractability in this formulation is on the belief that must capture the distribution of environments. Instead of directly operating on this distribution, the authors propose to learn a collision probability based on hand-coded features, while by striking out artificial features, our method is expected to achieve more flexible pattern recognition by jointly training perception and action modules. 

As mentioned in section \ref{sec:intro}, a work \cite{yu2024mavrl} released within weeks of our submission coincidentally studies speed-adaptive flight, but follows a different idea by training a policy directly outputting acceleration commands. In this approach, speed constraint is not explicitly incorporated into the action space. A specialized latent space used to maintain historical information is the key for the policy, which is achieved by a model-based mapping algorithm \cite{moravec1985high} in our framework. Our approach is tested in clutter (see Fig. \ref{fig:head}, Fig. \ref{fig:traj}, and Fig. \ref{fig:dense}) that are much more complex compared to those in \cite{yu2024mavrl}. The unique success of our approach stems not only from the learned policy, but also the generalizability and safety of the model-based trajectory planner \cite{zhou2020ego}. \vspace{-0.3cm}

\subsection{Combining Learning and Model-based Planner for Navigation in Cluttered Environments}

Recently, some works enhance traditional planning and control frameworks with learned modules to achieve more robust or efficient performance.  Previous works \cite{bansal2020combining,loquercio2021learning} employ imitation learning to directly predict local waypoints from visual input without geometric mapping for perceptive navigation. RL is also used to find an optimal policy \cite{faust2018prm} or a value estimator \cite{shah2022offline} as additional modules for navigation without manual labeling. The most similar of these works to ours in terms of technical pipelines are \cite{dobrevski2020adaptive,xu2021applr}, which employ RL to regularize hyperparameters for a classical local planning algorithm, Dynamic Window Approach (DWA). In these works, collision penalties can be densified by relaxing them as distance to the nearby obstacles. However, in this work, the spatial distribution of trajectories generated by the planner is almost independent of the output of the learned policy, so penalizing distance-related metrics is not causally meaningful, for which we employ a two-stage reward scheme. Moreover, the vehicle in our setup also has to deal with the limited field of view (FOV), rather than the near omnidirectional perception in \cite{dobrevski2020adaptive,xu2021applr}, which requires it to behave in a perception-aware manner. We explicit enable this by designing a local map representation (see section \ref{sec:5b}). While the above works are about ground robots navigating with low speed, our work contributes to agile flight systems, which requires the RL policy to cooperate with 3D perception and a much more complex modern planner backbone. Therefore, our work has a unique application contribution. \vspace{-0.1cm}

\section{Problem Formulation}
\label{sec:3}
We wish to control a flight vehicle through a prior unknown environment from one point to another with onboard perception. This problem can be formalized as a control problem in a partially observable Markov decision process (POMDP). We first overview the POMDP tuple $(S,A,T,R,\varOmega,O)$ and then formulate the problem.

\begin{itemize}
	\item States $S$. The states space is divided into two parts, one is the controllable states $S^c$ which consists of the states of the vehicle. The other part of $S$ is the environment $S^e$. In the context of our problem, $S^e$ can be described as the occupancy of the space, i.e., $s^e=\{0,1\}^n$.
	\item Actions $A$. Actions can be thought of as control commands in a general sense, which can be outputs of the trajectory planner or the actuator of the vehicle. 
	\item Conditional transition probabilities $T$. According to the definition of $S$, the transition probability is also divided into two independent components, where we impose no assumption to the state transition of $s^c$, while assuming $p(s^e|s^e,a_t)=1$, i.e., the environment $S^e$ is static.
	\item The reward function $R$. As the topic of this paper suggests, our reward function is defined as \emph{maximizing movement speed and minimizing collision loss}.
	\item Observations $\varOmega$. Observations include the state estimates of the vehicle, which is assumed to be perfect, and the partial observation of the environment $s^e$. The latter is limited by the FOV of the external sensor (in this paper, the depth camera) and occlusion by obstacles.  
	\item Conditional observation probabilities $O$. We preprocess the raw sensor input into a local occupancy map \cite{moravec1985high}. Thus, we are viewed as following the same assumptions about the conditional observation equations as \cite{moravec1985high}.
\end{itemize}

The action at each moment can be viewed as generated by a policy conditioned on the current observation and goal of navigation $g\in G$, i.e., $a_t\sim\pi \left( \cdot |o_t,g \right)$. The goal is to find the policy that maximizes the expected cumulative return,
\begin{equation}
	\label{eq:RL obj}
	\hspace{-0.3cm}\max_{\pi} \mathbb{E} _{s_{t+1}\sim p\left( \cdot |s_t,a_t \right) ,a_t\sim \pi \left(  \cdot |o_t,g \right),g\sim p_g}\hspace{-0.1cm}\left[ \sum\nolimits_t^{}{\gamma ^tr\left( s_t,a_t,g \right)} \right]\hspace{-0.1cm},
\end{equation}
where $p_g$ is the distribution of $g$, and $\gamma$ is the discount factor. \vspace{-0.6cm}

\section{Framework}
Our framework involves a learned policy, which is donated as $\pi ^{\dagger}$, a model-based trajectory planner, and the standard perception and low-level control ends. The policy $\pi ^{\dagger}$ and model-based trajectory planner form a hierarchical policy to optimize (\ref{eq:RL obj}). \vspace{-0.2cm}
\subsection{Model-based Trajectory Planner}
\label{sec:4A}
The trajectory planner receives the local geometric map and the states of the vehicle as inputs and generate a collision-free trajectory. The planner generates the trajectory by solving a constrained optimization in the form of \vspace{-0.1cm}
\begin{subequations}
	\begin{align}
		&\min_{u\left( t \right), T} \int_0^T{J\left( u\left( t \right) \right) +\rho \left( T \right)}, \\
		&\textnormal{s.t.} \nonumber \\
		&u\left( t \right) \in \mathcal{F} ,\ \ \forall t\in \left[ 0,T \right] , \\
		&\mathcal{G} \left( u\left( t \right) \right) \preceq \boldsymbol{0},\ \ \forall t\in \left[ 0,T \right], \label{eq:inequality} \\
		&\mathcal{H} \left( u\left( t \right) \right) =\boldsymbol{0}, \ \ \forall t\in \left[ 0,T \right], 
		\vspace{-0.1cm}
	\end{align}
\end{subequations}
where $u\left( t \right)$ is the planning quantities, $\rho :\left[ \left. 0,\infty \right) \right. \mapsto \left[ 0,\infty \right] $ the time regularization term, $\mathcal{F}$ the collision free region, $\mathcal{G} \left( \cdot \right) $ the dynamic constraints imposed by the physical limitation of the vehicle or users, and $\mathcal{H} \left( \cdot \right)$ summarizes the equality relations between the planning quantities and the states or control commands \cite{bertsekas2012dynamic}.

In the problem of navigation in cluttered environments, an inequality in (\ref{eq:inequality}) represents the level of aggressiveness on speed of vehicle $\boldsymbol{v}$, such as
$\left\| \boldsymbol{v} \right\| \leq \bar{v}$ or $\boldsymbol{v} \preceq \bar{\boldsymbol{v}}$  \cite{zhou2019robust,zhou2020ego}.  For the generality of our description, we write such constraints uniformly as 
\begin{equation}
	h\left( \boldsymbol{v} \right) \preceq v^{\dagger}
\end{equation}
without assigning them specific forms, where $v^{\dagger}$ is pre-determined as the planner solving for a trajectory.  \vspace{-0.1cm}

\begin{figure} \centering 
	\hspace{-0.3cm}  
	\includegraphics[width=0.95\columnwidth]{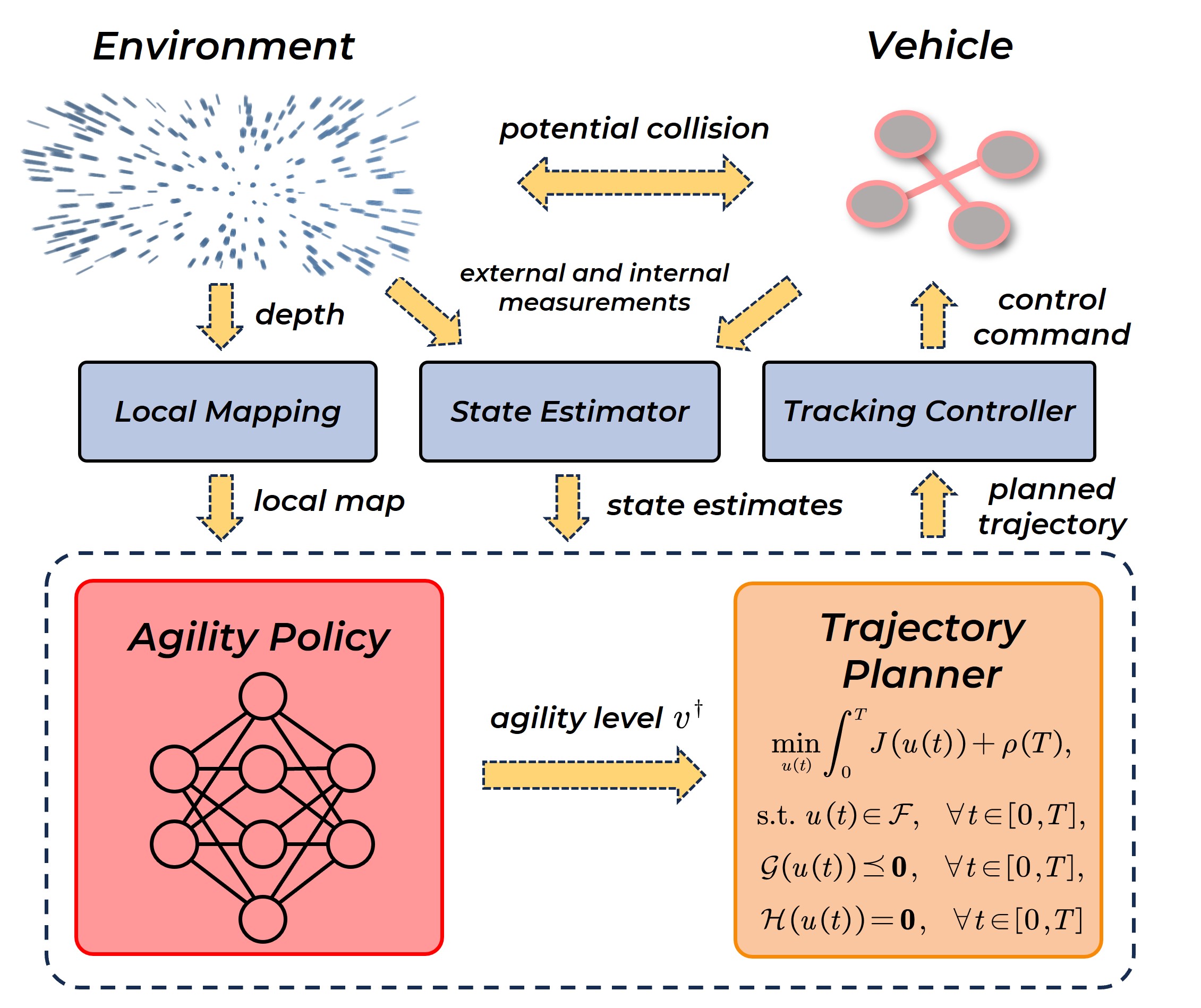} 
	\captionsetup{skip=5pt} 
	\captionsetup{font=footnotesize}  
	\caption{\textbf{Overview of the system with the hierarchical policy.}}
	\label{fig:system}   
	\vspace{-0.2cm} 
\end{figure}

\subsection{Hierarchical Policy Optimization}
Traditionally, $v^{\dagger}$ is conservatively determined in advance based on the user's observations of the environment to ensure safety. However, such an approach does not reasonably take advantage of the performance of the hardware and algorithmic systems. Instead, jointly optimizing the speed constraint and trajectory to optimize (\ref{eq:RL obj}) can further unlock the potential of the system.

We define a decomposition of the policy $\pi^\prime$ which augments $\pi$ in (\ref{eq:RL obj}) with an auxiliary action $v^{\dagger}$ \vspace{-0.5cm}

\begin{equation}
	\hspace{-0.2cm}\pi^\prime( a^\prime|o,g ) \coloneqq \pi^\prime( a,v^{\dagger}|o,g ) =\pi ''( a|v^{\dagger},o,g ) \cdot \pi ^{\dagger}( v^{\dagger}|o,g ), 
\end{equation}
where  $\pi^\prime$ is decomposed as two hierarchical policies $\pi ''$ and $\pi ^{\dagger}$, the former of which is conditioned on the auxiliary action while the latter decides it. The action space of the original POMDP is augmented as a result of the introduction of an auxiliary action, i.e., 
$A^\prime=A\times V^{\dagger}$. However, no other part of the POMDP is changed, so we can define a \emph{hierarchical policy optimization} problem
\begin{equation}
	\label{eq:RL obj2}
	\begin{split}
		\max_{\pi '',\pi ^{\dagger}} \mathbb{E} _{v^{\dagger}_t\sim \pi ^{\dagger}\left( \cdot |o_t,g \right), a_t\sim \pi ''( \cdot |v^{\dagger}_t,o_t,g )}  \left[ \sum\nolimits_t^{}{\gamma ^tr\left( s_t,a_t,g \right)} \right]=\\ \max_{\pi ^{\dagger}}\max_{\pi ''}\mathbb{E} _{v^{\dagger}_t\sim \pi ^{\dagger}( \cdot |o_t,g ), a_t\sim \pi ''( \cdot |v^{\dagger}_t,o_t,g )}\left[ \sum\nolimits_t^{}{\gamma ^tr( s_t,a_t,g )} \right],
	\end{split}
\end{equation}
equivalent to (\ref{eq:RL obj}), where, for more concise writing, we omit the state and goal distributions in (\ref{eq:RL obj}). This problem is equivalent to (\ref{eq:RL obj}) in the sense that if $\pi$ and $\pi^\prime$ impose the same distribution on $a$ when conditioned on any  $o\in O$ and $g\in G$, the objective function values of (\ref{eq:RL obj}) and (\ref{eq:RL obj2}) are consistent.

Model-based trajectory planners \cite{loquercio2021learning,wang2022geometrically,zhou2019robust,zhou2020ego}, after decades of development, \emph{can be regarded as a good approximation of the optimal} $\pi''$ conditioned on a given $v^{\dagger}$. Therefore, only the outer-loop policy $\pi ^{\dagger}$ needs to be learned. We use online model-free RL to learn this policy, which is detailed in the next section.   \vspace{-0.1cm}

\subsection{System Overview}
We conclude this section with an overview of the whole system, as illustrated in Fig. \ref{fig:system}.

At the uppermost end of the system, the vehicle obtains partial observations of the environment via external sensors, which are used to provide state feedback (e.g., via the visual-inertial odometry (VIO)) and to construct local occupancy maps (e.g., from the depth camera). The state estimates and local map are fed to the  policy $\pi ^{\dagger}$ and trajectory planner. We choose to use the occupancy map as the input of the RL policy instead of the raw sensory data because the training is performed in simulation, and such an approach deals with sensor noise through a model-based approach, i.e., the employed mapping algorithm, thus bridging the gap between simulation and reality in RL training.

At each time step $t$, the policy $\pi ^{\dagger}$ outputs a high-level planning instruction $v ^{\dagger}_t$. The trajectory planner generates a trajectory  $u$ in a horizon $[t,t+T]$ conditioned on $v ^{\dagger}_t$, which is treated as the action $a_t$ of the POMDP. The trajectory is tracked by a low-level controller and executed by the vehicle's inner-loop controller. In our definition, the unknown environment as well as  tracking and actuating controllers are treated as dynamics of the POMDP.  \vspace{-0.1cm}

\begin{figure} \centering 
	\hspace{-0.3cm}  
	\includegraphics[width=0.9\columnwidth]{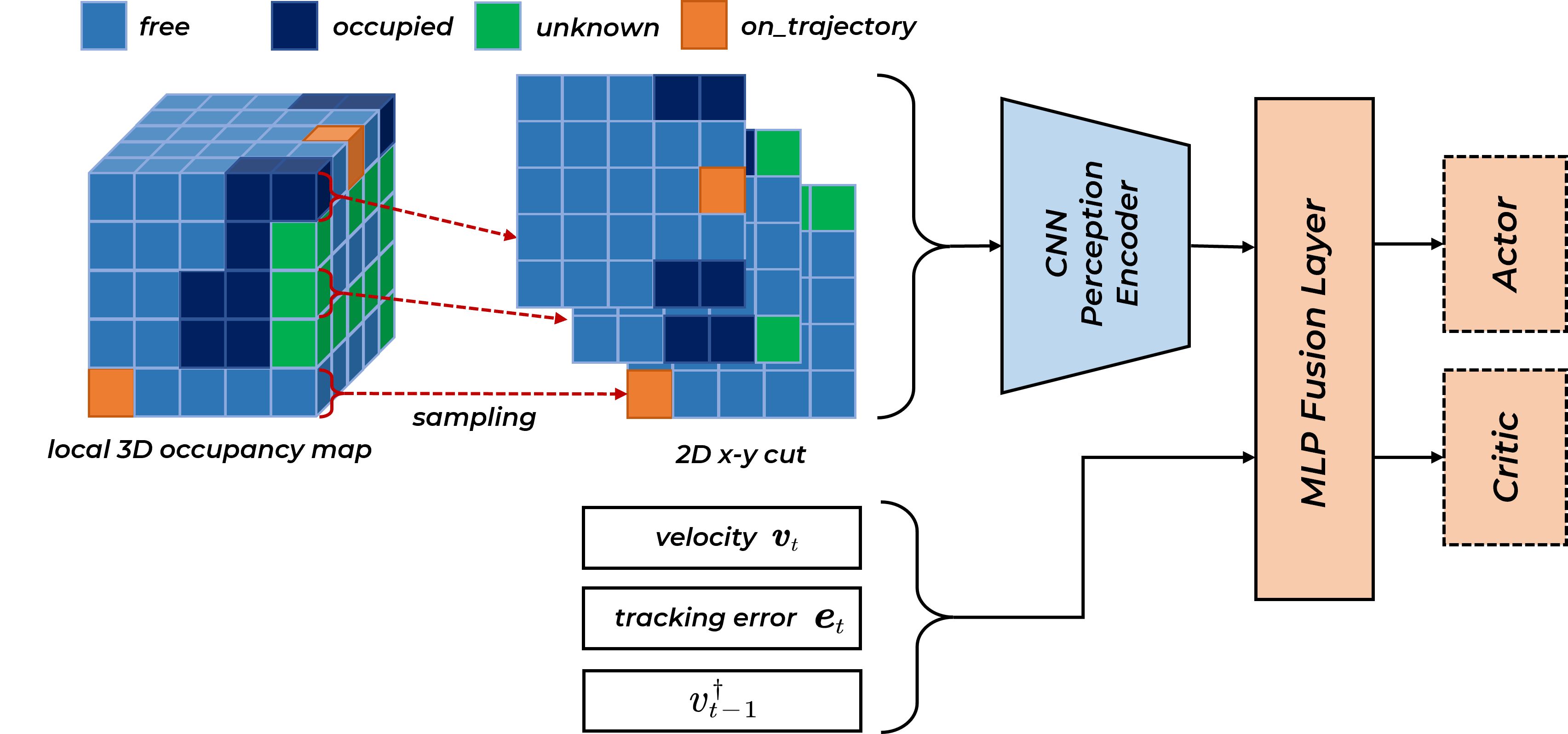} 
	\captionsetup{skip=5pt} 
	\captionsetup{font=footnotesize}  
	\caption{\textbf{Illustration of the policy architecture and observation implementation.} The figure shows the network architecture of policy, where actor and critic share the CNN encoder and fusion layer. We also highlight the 3D occupancy map where each cell is assigned a state, and its x-y profiles are sampled as the input of the network.}
	\vspace{-0.2cm}
	\label{fig:network}    
\end{figure}

\section{Reinforcement Learning for the Outer-loop Policy}

In this section, we detail how to effectively learn the outer-loop policy. Eq. (\ref{eq:RL obj2}) implies an approach to learn this policy, which treats also the trajectory planner as part of the environment dynamics. An auxiliary POMDP is considered to learn $\pi ^{\dagger}$, in which the action space is $V^{\dagger}$. In the following subsections, we present the implementation of each part of the RL policy.
\vspace{-0.15cm}
\subsection{Policy Representation}

We represent policy and value functions using a branch of partially shared neural networks, as illustrated in Fig. \ref{fig:network}. The networks share a perception encoder consists of convolution neural networks (CNNs) and a fusion layer implemented with multilayer perceptron (MLP). We implement the CNN encoder with 4 layers whose [channel number, kernel size, stride, padding] are [5,5,3,2], [32,3,2,1], [48,3,2,1] and [64,3,1,0]. The result of the CNN encoder is flattened and concatenated into the
inputs of the MLP fusion layer. The output dimension of the MLP layer is 64. The actor and critic networks are both implemented as MLPs with 2 hidden layers having 64 and 32 units, respectively.

\vspace{-0.15cm}
\subsection{Observation Space}
\label{sec:5b}
\vspace{-0.15cm}
As implied by (\ref{eq:RL obj2}), in principle, $\pi ^{\dagger}$ should be conditioned on the observations and the goal of navigation. However, for more effective learning, we design a specialized input space. As illustrated in Fig. \ref{fig:network}, the input consists of the 3D local occupancy map $\boldsymbol{m}_t$, the current velocity $\boldsymbol{v}_t$, the current tracking error $\boldsymbol{e}_t\coloneqq \boldsymbol{p}_t-\hat{\boldsymbol{p}}_t$, where $\boldsymbol{p}_t$ and $\hat{\boldsymbol{p}}_t$ are the true and desired position of the vehicle, respectively, the decision at the last time step ${v}^{\dagger}_{t-1}$, and a \emph{pre-planned} trajectory at the current time step generated according to ${v}^{\dagger}_{t-1}$ and $o_t$, which will not be executed by the controller.     

The pre-planned trajectory serves two purposes. First, it expresses the current position of the vehicle in the local map as well as the local target chosen by the trajectory planner according to the global goal $g$. Second, since the speed component of the trajectory does not have a significant effect on the spatial distribution of the trajectory, we can use this pre-planned trajectory as a spatial approximation of the trajectory to be planned, and therefore alert the perceptual part of the policy as to which regions in the local map are important.

We correlate the trajectory and the 3D local map by 'drawing' the former on the latter, as illustrated by Fig. \ref{fig:network}. Specifically, each cell in the occupancy map is assigned one of four states: \texttt{free}, \texttt{occupied}, \texttt{unknown}, and \texttt{on\_trajectory}. The state \texttt{unknown} is set to trigger perception-aware behaviors. For more efficient and easier learning, we sample the trajectory with time interval $\delta t$, keeping only the x-y 2D cuts of the 3D map corresponding to the sampled points on the trajectory as input to the network, which is illustrated in Fig. \ref{fig:network}. \vspace{-0.15cm}

\subsection{Early Termination}

\label{sec:5C}
We define two cases of termination. One is when the vehicle is in an unsafe state, i.e., the vehicle is judged in the \texttt{emergency\_stop} or   \texttt{collided} state, which is determined by the state machine in a modern motion planning system \cite{zhou2019robust,zhou2020ego}. The other occurs when the planning solution procedure does not finish in a valid time, which is often due to too high level of aggressiveness allowed in the previous time steps, so that the vehicle is too close to a suddenly appearing obstacle under the perception latency and thus the planner fails to plan a feasible trajectory.

\vspace{-0.15cm}
\subsection{Reward Function}
\label{sec:reward}

 As described in section \ref{sec:3}, we design the reward function to maximize traversal speed and minimize collision loss. For practical considerations, we further divide the reward function into four parts: the speed term, as follows:
\begin{equation}
	r=r_{\textnormal{speed}}+r_{\textnormal{smoothing}}+r_{\textnormal{error}}+r_{\textnormal{danger}}, \label{eq:rwd}
\end{equation}
where the specific forms of the smoothing, tracking error and collision terms are
\begin{equation}
	r_{\textnormal{smoothing}} = -\lambda_{\textnormal{smoothing}}
	\| v_{t}^{\dagger}-v_{t-1}^{\dagger} \| 
	^2, \label{eq:rwddv}
\end{equation}
\begin{equation}
	r_{\textnormal{error}}=-\lambda _{\textnormal{error}}\min \left\{ \|\boldsymbol{e}_t\| ,e_{\max} \right\} ^2,
\end{equation}
\begin{equation}
	r_{\textnormal{danger}} =
	\begin{cases} 
		-\lambda_{\textnormal{danger}}\|\boldsymbol{v}_t\|^2 & \textnormal{(when the episode is terminal)}\\
		0 &\textnormal{(otherwise)}\\
	\end{cases},
\end{equation}
for some $\lambda_{\textnormal{smoothing}}>0$,  $\lambda_{\textnormal{error}}>0$, and $\lambda_{\textnormal{danger}}>0$. 

However, $r_{\textnormal{danger}}$ is highly sparse and stochastic, where the stochasticity is derived from the complex modern trajectory planner. This distribution of reward can cause difficulties when learning the policy with general RL algorithms. One way to relax the early-termination penalty and make it denser is to add the distance to obstacles into the reward function. However, although this may be effective for an end-to-end policy that directly outputs a low-level command \cite{yu2024mavrl}, this kind of geometric metric cannot evaluate the risk of early termination in our framework, as the spatial distribution of the vehicle is mainly determined by the planned trajectory where smaller distances to obstacles do not necessarily imply that collisions are more likely to occur \cite{zhou2020ego}. In principle, we should only settle for penalizing as an episode does terminate, which is causally relevant to some implicit factors such as perception latency and tracking failure that make overly aggressive trajectories prone to cause collisions.

Our key finding here is that with a two-stage reward, learning can take place efficiently and effectively. The only term that differs between the first and second stages is $r_{\textnormal{speed}}$, which is detailed in the following.

\subsubsection{A human knowledge-based, dense reward function for pre-training} 
The speed reward in the first stage, where $v_{t}^{\dagger}$ is imposed to be larger or equal to a certain constant value when training (e.g., 1m/s), is defined as follows:
\begin{equation}
	\label{eq:stage1}
	r_{\textnormal{speed}} =
	\begin{cases} 
		\lambda_{\textnormal{speed}}^1(\phi _t^1 - \|\boldsymbol{v}_t\|) &(\phi _t^2 > \lambda_{\phi}^1)\\
		\lambda_{\textnormal{speed}}^2( \|\boldsymbol{v}_t\| - \phi _t^1) &(\phi _t^2 < \lambda_{\phi}^2)\\
		\lambda_{\textnormal{speed}}^3 \|\boldsymbol{v}_t\|& \textnormal{(otherwise)}\\
	\end{cases},
\end{equation}
where $\lambda_{\textnormal{speed}}^i>0$ for $i\in \left\{ 1,2,3 \right\}$ and $\lambda_{\phi}^i$ for $i\in \left\{ 1,2 \right\} $ are hyperparameters such that $\lambda_{\textnormal{speed}}^2 > \lambda_{\textnormal{speed}}^3$, and $\phi _t^i$s, for $i\in \left\{ 1,2 \right\} $, are handcrafted feature values encoding the obstacle distribution. Specially, $\phi _t^i$s are normalized linear combinations of the distance to the nearest obstacle, volume of the obstacle, and number of obstacles at current time . 

In (\ref{eq:stage1}), $\phi _t^1$ is designed to shape the reward function smoother with a zero mean and $\phi _t^2$ is a naive estimate of the level of danger by human knowledge. Therefore, the first case in (\ref{eq:stage1}) serves as a \emph{dense smoothing} of the collision penalty, which imposes the policy to behave conservatively when the observed environment is manually evaluated as dangerous. Such an approach prevents the policy from being single-mindedly greedy to gain instant reward by accelerating. Note that we do not necessarily carefully tune the hyperparameters in (\ref{eq:stage1}), but leave the task of further optimization to the objective reward in the second stage.

\subsubsection{An objective reward for fine-tuning}
In the second stage, we restore the original intent of the speed reward, which is only to encourage greater aggressiveness:
\begin{equation}
	\label{eq:stage2}
	r_{\textnormal{speed}} = \lambda_{\textnormal{speed}}^3 \|\boldsymbol{v}_t\|. 
\end{equation}
\vspace{-0.3cm}

\section{Training Setup and Implementation}

\subsection{System Setup}

\label{sec:5A}
We choose a state-of-the-art model-based trajectory planner, EGO-planner \cite{zhou2020ego}, as the backbone. The speed constraint is imposed in the form of $\left\| \boldsymbol{v} \right\| \leq \bar{v}$. The local mapping algorithm is implemented based on \cite{moravec1985high}. The vehicle is equipped with a forward-facing camera that can measure depth with a FOV of 87$^\circ$ (horizontal) $\times$ 58$^\circ$ (vertical) and a valid/trusted depth range from $0.28$m to $5$m. The refreshing frequency of  
 the depth data is 15 Hz. A PID controller running at 50 Hz is used for position tracking.

The outer-loop policy runs at 10 Hz. Conceptually, trajectory planning is required after each outer-loop policy output, but too frequent replanning increases the situations of triggering the \texttt{emergency\_stop} state in practice, as well as increasing the computational overhead. Therefore, we design a criterion to determine whether replanning is imposed by the outer-loop policy output, and if this criterion is not met, the planner follows its original implementation of replanning rules \cite{zhou2020ego}. In particular, if $v_{t}^{\dagger}-v_{t-1}^{\dagger}\in \left[ -0.3,0.5 \right]$m/s, we do not impose a replanning of the trajectory planner.
\vspace{-0.15cm}
\subsection{Training Environment}
\label{sec:factors}

We train the policy in a customized simulator, which parallelizes multiple (30 in our implementation) separate environments. Unlike general simulation environments, the simulator does \emph{not} stop the clock while solving for the action, i.e., mapping and trajectory planning, in our simulation environment. Meanwhile, the sensory data is updated \emph{asynchronously} with the action. These specialized designs are elaborated to simulate one of the most important factors limiting allowed aggressiveness, the perception latency, i.e., the time interval between perception and execution of an action. Even with these designs, however, we can only approximate perception latency because the computing platform at the deployment time is different from the one at the time of training.  Another crucial factor that limits aggressiveness, the error of trajectory tracking due to limited performance of controller or possible dynamically infeasible trajectories, is also embedded in the environment by setting up the actual PID tracking controller for the vehicle, as described earlier.

\begin{figure} \centering 
	\hspace{-0.3cm}  
	\includegraphics[width=1.0\columnwidth]{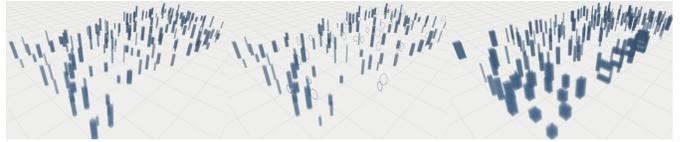} 
	\captionsetup{skip=5pt} 
	\captionsetup{font=footnotesize}  
	\caption{\textbf{Example illustration of environments for training.} Areas in blue represent the space filled with obstacles.}
	\vspace{-0.3cm}
	\label{fig:envs}    
\end{figure}

\subsection{Training Implementation}

The policy is learned with the soft actor-critic (SAC) algorithm \cite{haarnoja2018soft}. Prioritized experience replay (PER) \cite{schaul2015prioritized} is also applied to mitigate the sparse failure mode problem in a general way. We train the policy in multiple scenarios with variable obstacle distributions, some of which are shown in Fig. \ref{fig:envs}. We \emph{freeze} the CNN encoder in the second stage of training, considering that the first-stage reward design encourages the encoder to extract features that can flexibly respond to different patterns of the obstacle distribution.  \vspace{-0.2cm}

\begin{figure*} \centering
	\subfloat {  
		\includegraphics[width=0.076\columnwidth]{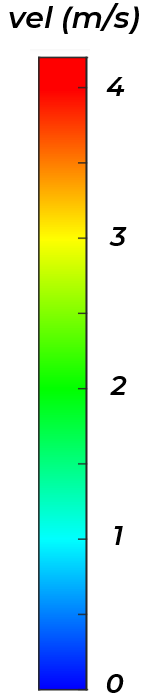} 
	}    
	\setcounter{subfigure}{0} 
	\subfloat[] {
		\label{fig:son}
		\includegraphics[width=0.58\columnwidth]{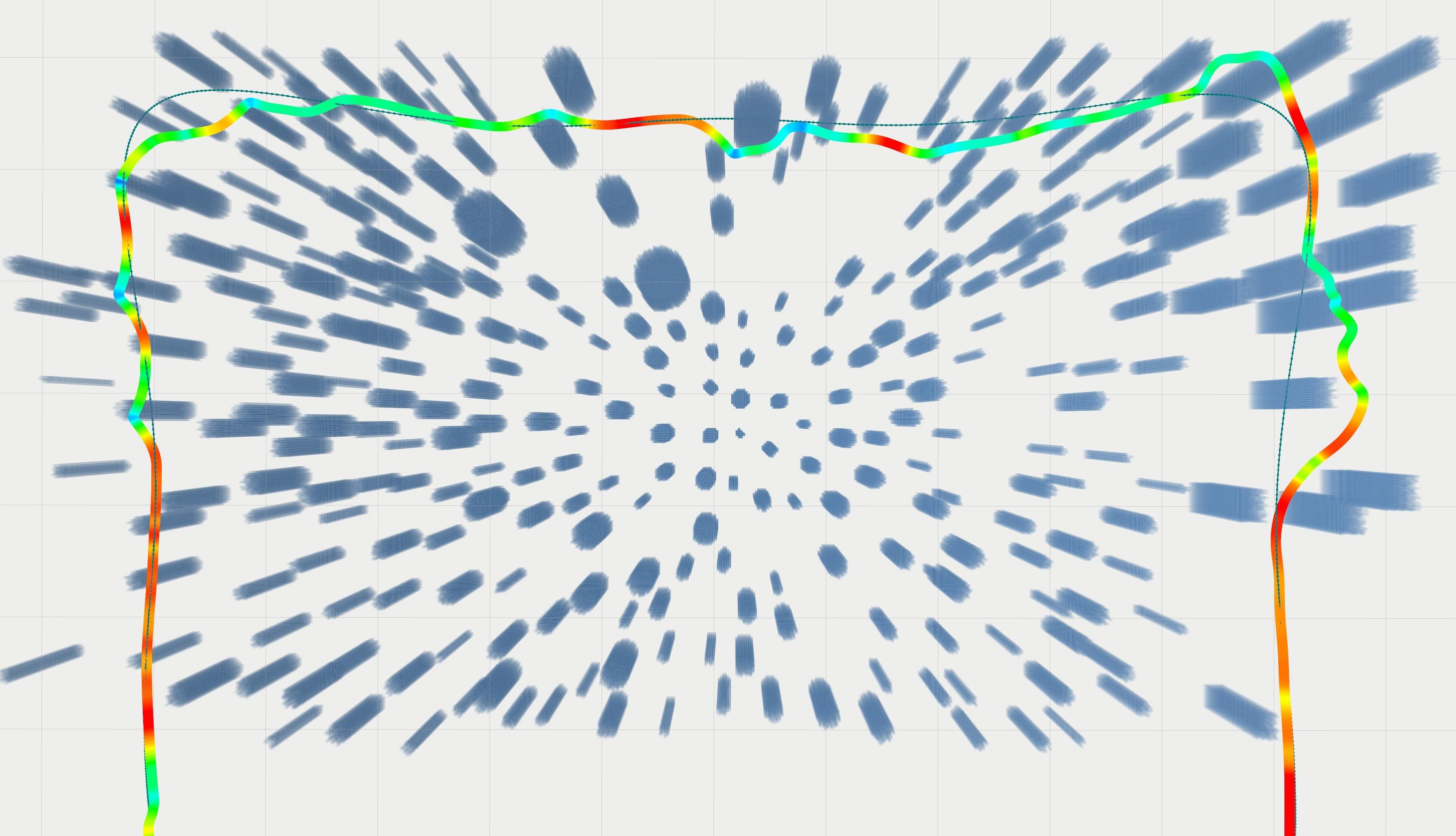}
	}
	\hspace{-0.1cm}\subfloat[] {    
		\label{fig:v2}
		\includegraphics[width=0.58 \columnwidth]{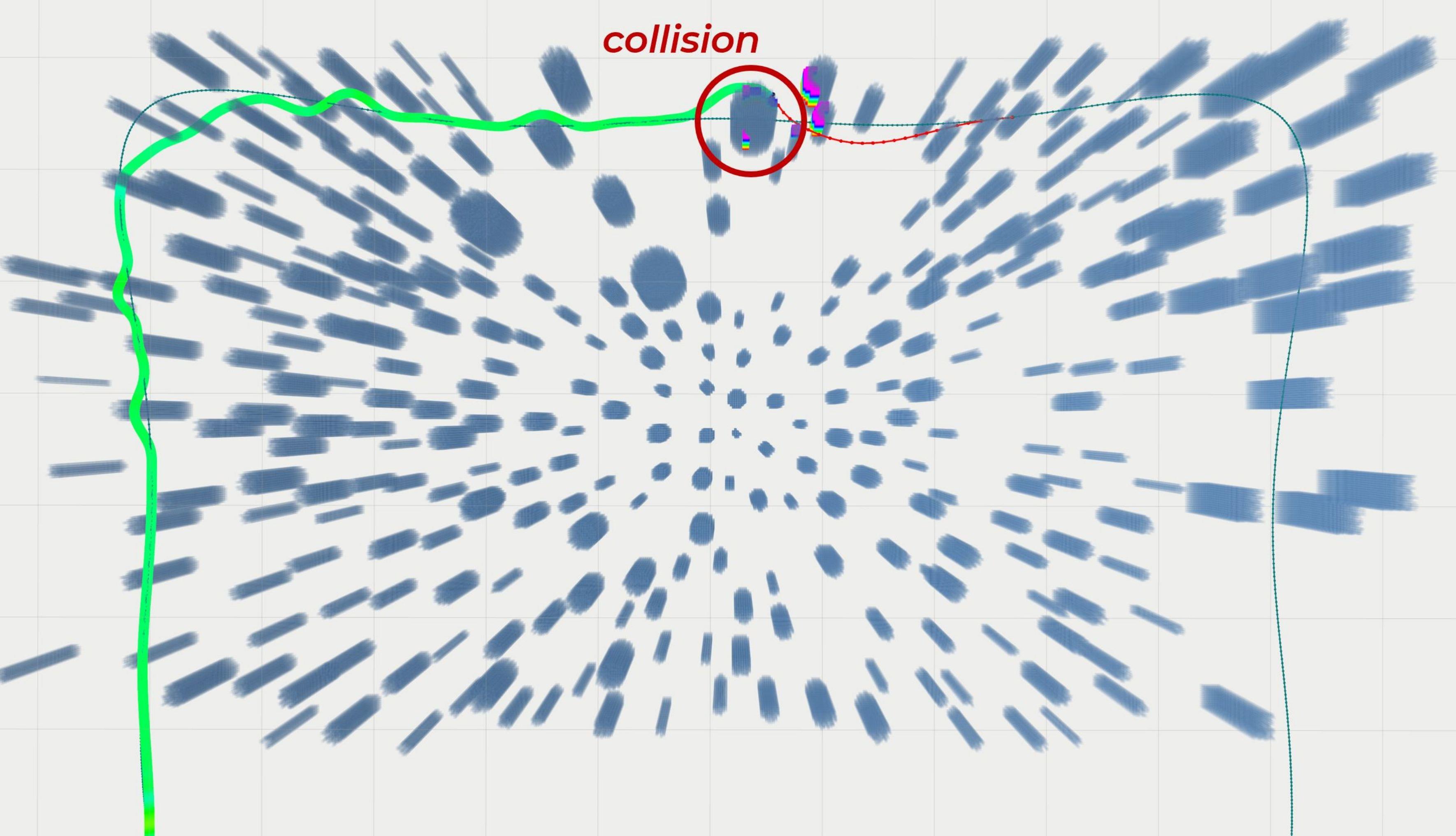}     
	}
	\hspace{-0.1cm}\subfloat[] {    
		\label{fig:father}
		\includegraphics[width=0.581\columnwidth]{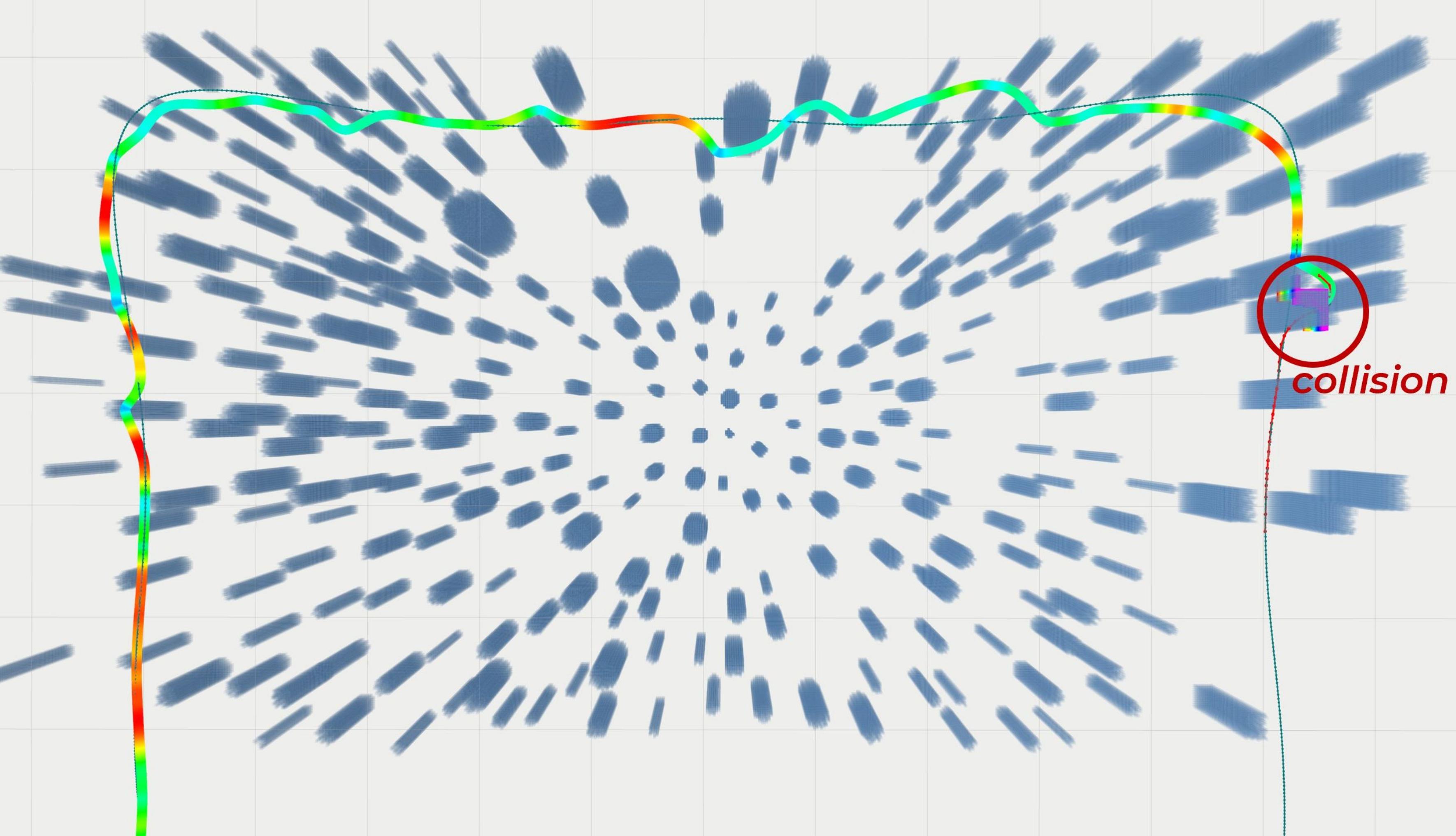}     
	}
\\  
	\captionsetup{font=footnotesize}
	\caption{ \textbf{Velocity distribution along the trajectory in different setups.} The dark green curves are the reference trajectory. The colorful curves are the trajectory that the vehicle passes over, where the color represents the velocity.  (a) An example result of the proposed approach. (b) An example result of the constant speed constraint ${v}^{\dagger}=2$m/s. (c) An example result of the intermediate model before fine-tuning.}
	\label{fig:traj}
	\vspace{-0.3cm}
\end{figure*}

\begin{figure} \centering 
	\hspace{-0.3cm}  
	\includegraphics[width=0.75\columnwidth]{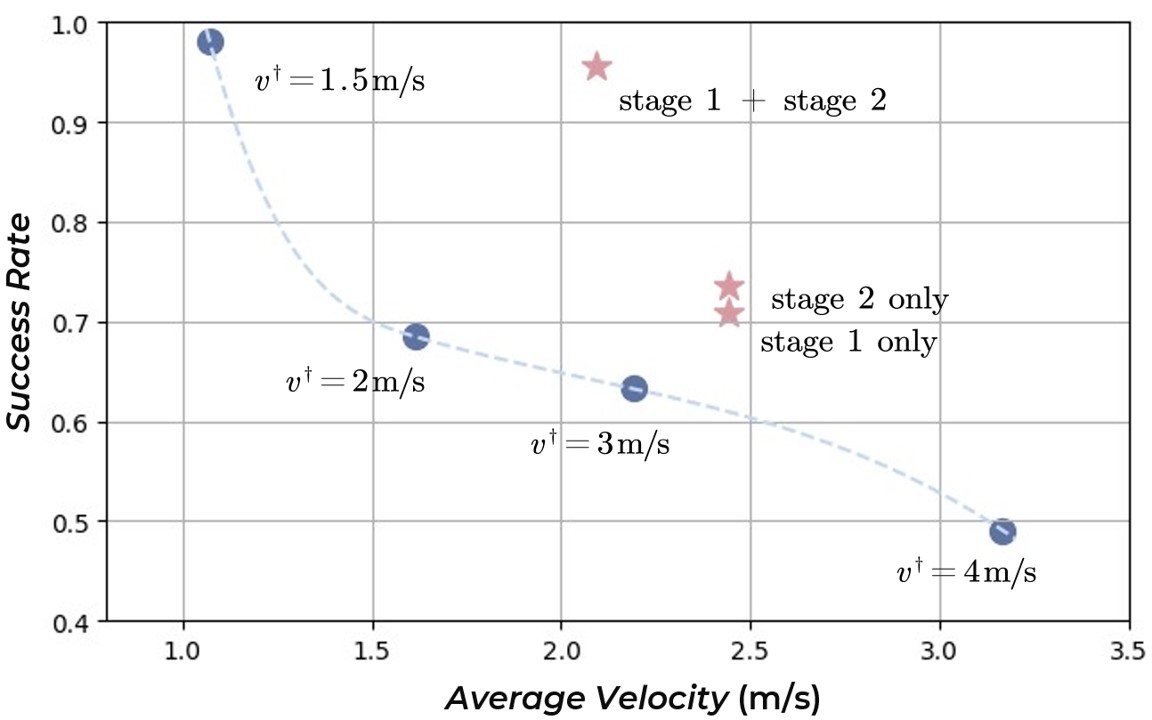} 
	\captionsetup{skip=5pt} 
	\captionsetup{font=footnotesize}  
	\caption{\textbf{Statistical results of different setups.} Success rates are computed on 50 trials for each setup. The light blue dashed lines connect the statistics at different levels of constant speed constraint, indicating the inherent capability of the system. Average velocities are computed as the average of success trials in the 50 trials.}
	\label{fig:statistics}  
	\vspace{-0.3cm}  
\end{figure}

\section{Experiments}

\subsection{Simulation Experiments}

We benchmark three approaches in the simulation experiments: (i) the proposed 'learned policy + model-based planner backbone', (ii) the backbone planner EGO-planner with constant speed constraint (i.e., constant ${v}^{\dagger}$), and (iii) the handcrafted cost function-based (non-learning) environmental adaptive planner EVA-planner \cite{quan2021eva}, where the default parameters in the cost function are used, and all parameters that EGO-planner shares with it are set equal to those of EGO-planner.  

\begin{figure} \centering
	
	\subfloat {  
		\includegraphics[width=0.048\columnwidth]{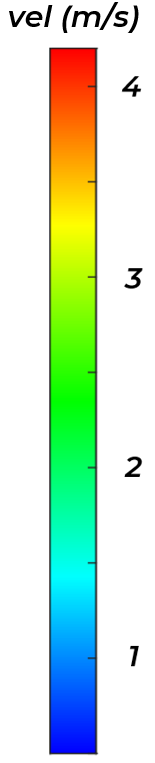}
	}  
	\setcounter{subfigure}{0} 
	\subfloat[] {
		\label{fig:APC}
		\includegraphics[width=0.42\columnwidth]{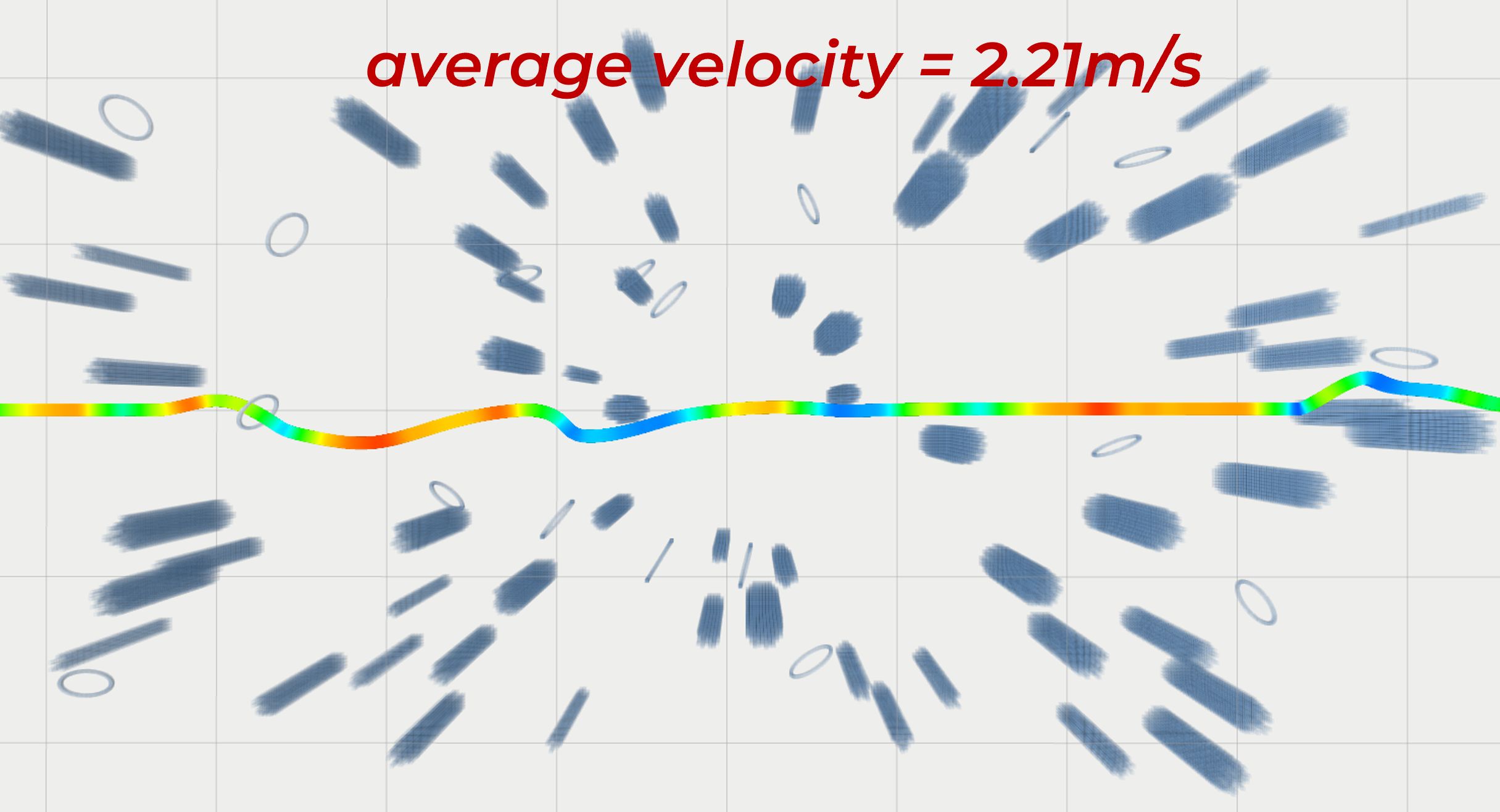}
	}
	\hspace{-0.1cm}\subfloat[] {    
		\label{fig:EVA}
		\includegraphics[width=0.42\columnwidth]{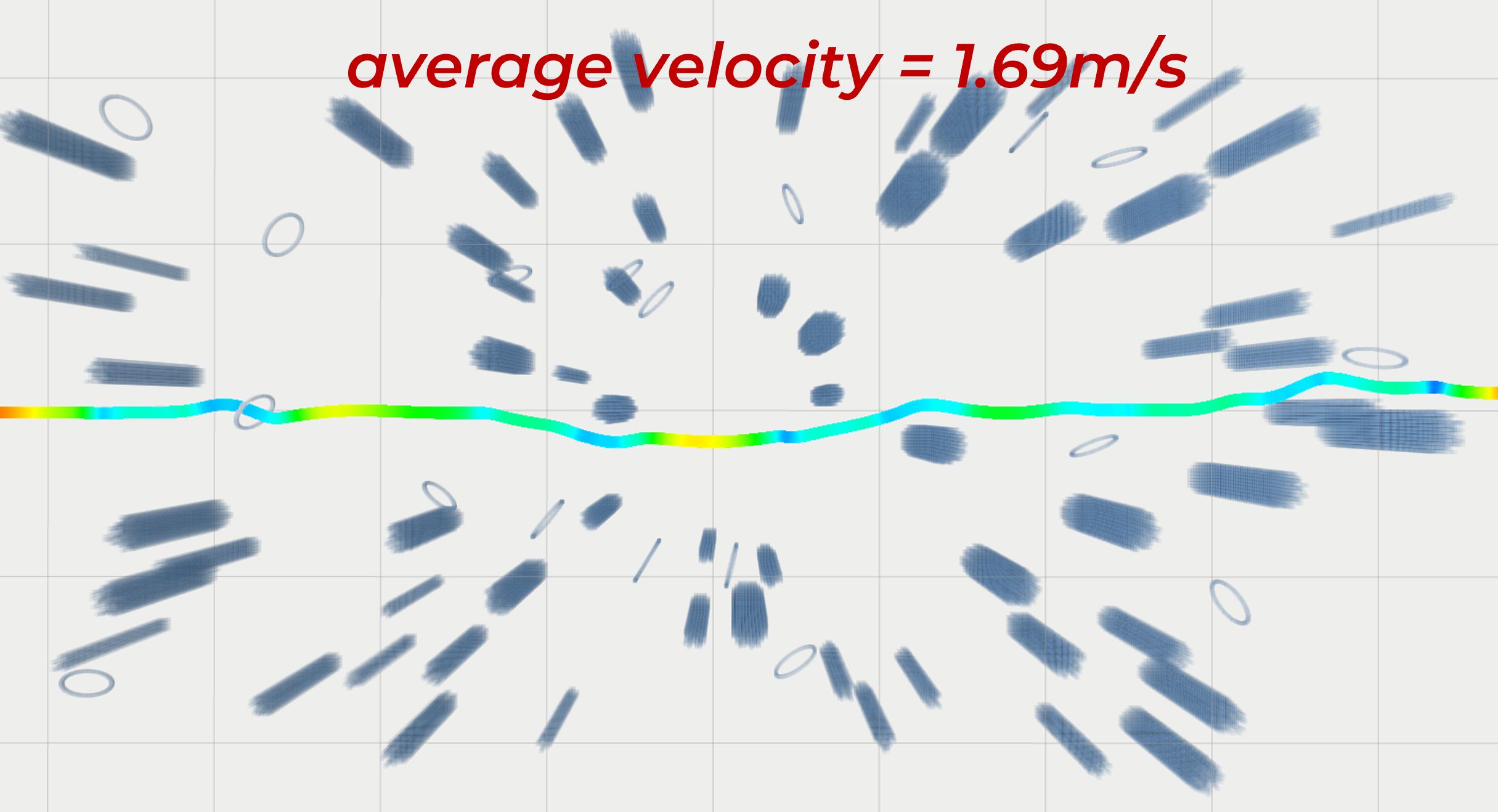}     
	}\\  
	\captionsetup{font=footnotesize}
	\caption{ \textbf{Comparison of the proposed approach and EVA-planner.} The colorful curves are the trajectory that the vehicle passes over, where the color represents the velocity. The marked average velocity is computed on 30 trials. (a) Illustration of the trajectory generated by the proposed approach. (b) Illustration of the trajectory generated by EVA-planner.}

	\label{fig:EVAvel}
	\vspace{-0.3cm}
\end{figure}

\begin{figure}[h] \centering
	\hspace{-0.2cm}
	\subfloat {  
		\includegraphics[width=0.05\columnwidth]{legend2.png}
	}  
	\setcounter{subfigure}{0} 
	\hspace{-0.2cm}\subfloat[] {
		\includegraphics[width=0.156\columnwidth]{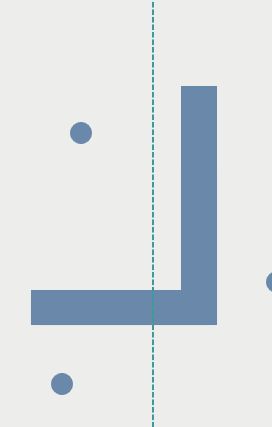}
	}
	\hspace{-0.2cm}\subfloat[] {
		\includegraphics[width=0.36\columnwidth]{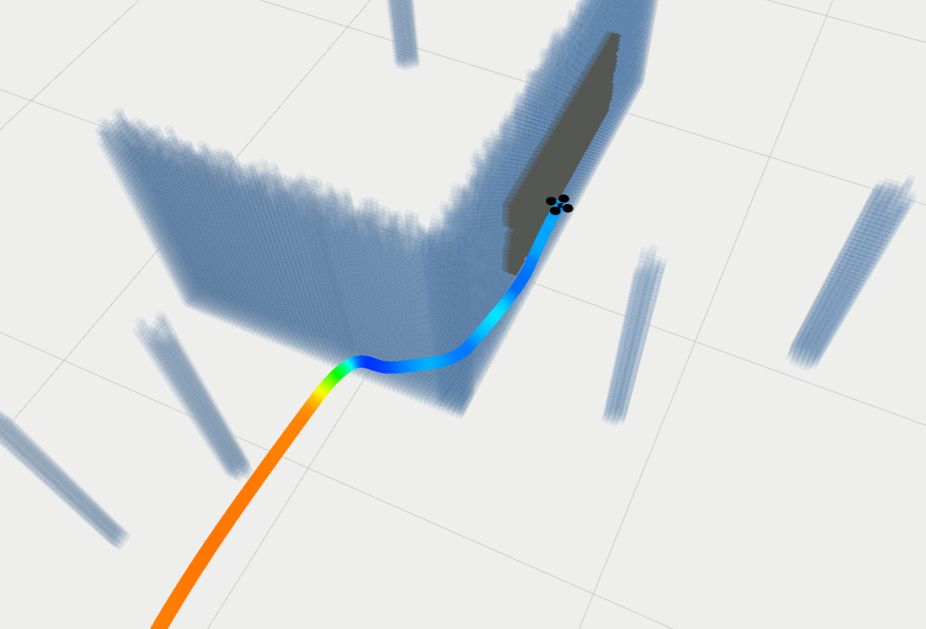}
	}      
	\hspace{-0.2cm}\subfloat[] {    
		\includegraphics[width=0.36\columnwidth]{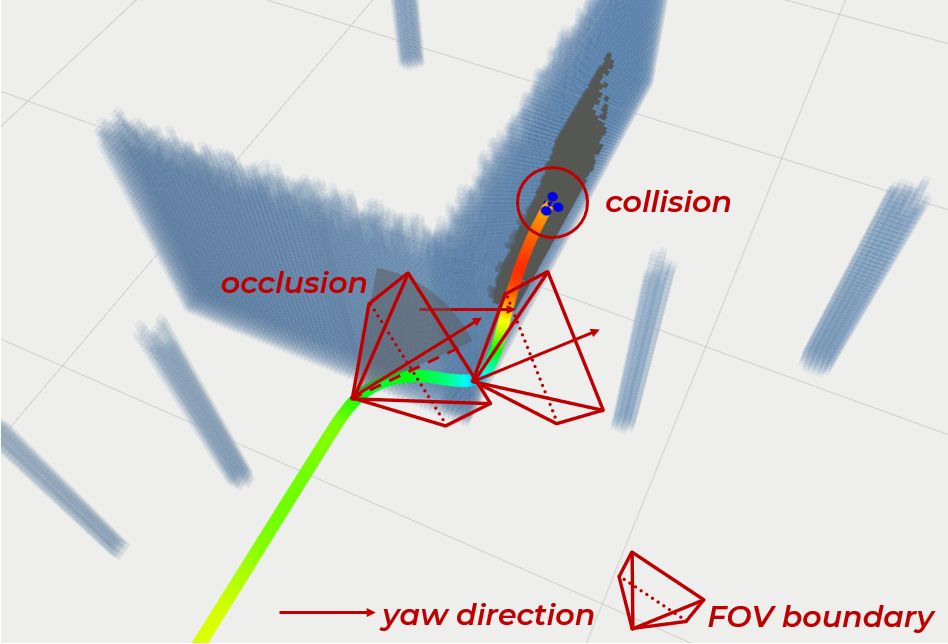}     
	} \\  
	\captionsetup{font=footnotesize}
	\caption{ \textbf{Illustration of perception-aware behaviors of the learned policy and comparison with EVA-planner.} The colorful curves are the trajectory that the vehicle passes over, where the color represents the velocity. The dark gray grid is the built local map. (a) Illustration of the scenario, where the dark green line is the reference trajectory. (b) Behaviors of the proposed approach. (c) Behaviors of EVA-planner, where the yaw angles and FOV boundaries are shown to explain the failure.}
	\vspace{-0.4cm}
	\label{fig:perception-aware}
\end{figure} 

\subsubsection{Comparison of the proposed framework with constant speed constraint baselines and ablation tests}
The first two setups are evaluated in a challenging environment as shown in Fig. \ref{fig:traj}, where obstacles of different densities, sizes, and appearances are unevenly distributed. The evaluation environment is never seen when training. In Fig. \ref{fig:statistics}, we report the statistical results of traversal velocity in success cases and the success rate which is defined as the proportion of trials that complete the entire track. A trial is judged
to be ’terminal’ with the same criteria as during training. In Fig. \ref{fig:traj}, we visualize the trajectories and velocities at every moment of the proposed approach and the baselines. Since the trajectory generation backbone of EVA-planner cannot always effectively generate feasible trajectory in such a complex environment, it is not fair to include it in this statistical result, so we ignore EVA-planner in Fig. \ref{fig:statistics}. 

The statistics show that as ${v}^{\dagger}$ is set higher, the success rate decreases, creating a sloping downward performance curve in Fig. \ref{fig:statistics}. Such performance curves capture the inherent performance of the integration of physical (e.g., sensors equipped) and algorithmic (e.g., the mapping, planning and control framework) systems. Although under the performance constraints of the system, by dynamically configuring the speed constriant, the vehicle combines the exploitation of agility and safety guarantees. It can be seen in Fig. \ref{fig:statistics} that the policy learned with both the reward stages achieves similar success rate as the lowest speed case where ${v}^{\dagger}=1.5$m/s, but also exhibits considerable traversal speeds to efficiently complete the track. In contrast, a relatively aggressive constant speed constraint ${v}^{\dagger}=2$m/s can frequently lead to emergency stops or collisions in areas with dense obstacles or local occlusion, as shown in Fig. \ref{fig:v2}.

 In Fig. \ref{fig:statistics}, we also show the performance of the policies trained only with either the first stage (reward (\ref{eq:stage1})) or the second stage (reward (\ref{eq:stage2})). The results  indicate that although the final policy can successfully improve the overall performance of the system, the policy trained with only the first stage does not exhibit such improvement. This result is not so surprising since the reward scheme of the first stage of training does not reflect the objective values and impose inaccurate human knowledge. On the other hand, the necessity of the first training stage is reflected by the inferiority of the performance rendered by the policy trained directly with sparse (but objective) rewards in Fig. \ref{fig:statistics}.


\subsubsection{Comparison of the proposed framework with EVA-planner and an example case of perception-aware behaviors}
EVA-planner tends to plan (sometimes overly) conservative velocities, which can be seen in Fig. \ref{fig:EVA}, while the proposed approach makes better use of the vehicle's maneuverability, at the cost of large changes in velocity, as seen in Fig. \ref{fig:APC}. 

More importantly, since the hand-designed cost function of EVA-planner does not capture some aspects of the problem, such as the necessity of perception awareness, an unsafe decision may be output by the planner. An example is illustrated in Fig. \ref{fig:perception-aware}. In such a case, the outer-loop policy learns to behave cautiously when the vehicle plans to turn its head into area out of its perception range, which is caused by occlusion of a corner and attitude angle changes. In contrast, EVA-planner plans an unsafe velocity due to its illusion that there are no obstacles nearby, which is caused by occlusion, yaw angle changes, and perception latency. However, such perception-aware behaviors does not always occur in the policy obtained using only the first stage of reward. An example is shown in Fig. \ref{fig:father}, where acceleration at a corner causes the vehicle to collide. \vspace{-0.2cm}

\begin{figure*} \centering
	\subfloat {  
		\includegraphics[width=0.067\columnwidth]{legend2.png}
	}  
	\setcounter{subfigure}{0} 
	\hspace{-0.1cm}\subfloat[] {
		\label{fig:2.5}
		\includegraphics[width=0.55\columnwidth]{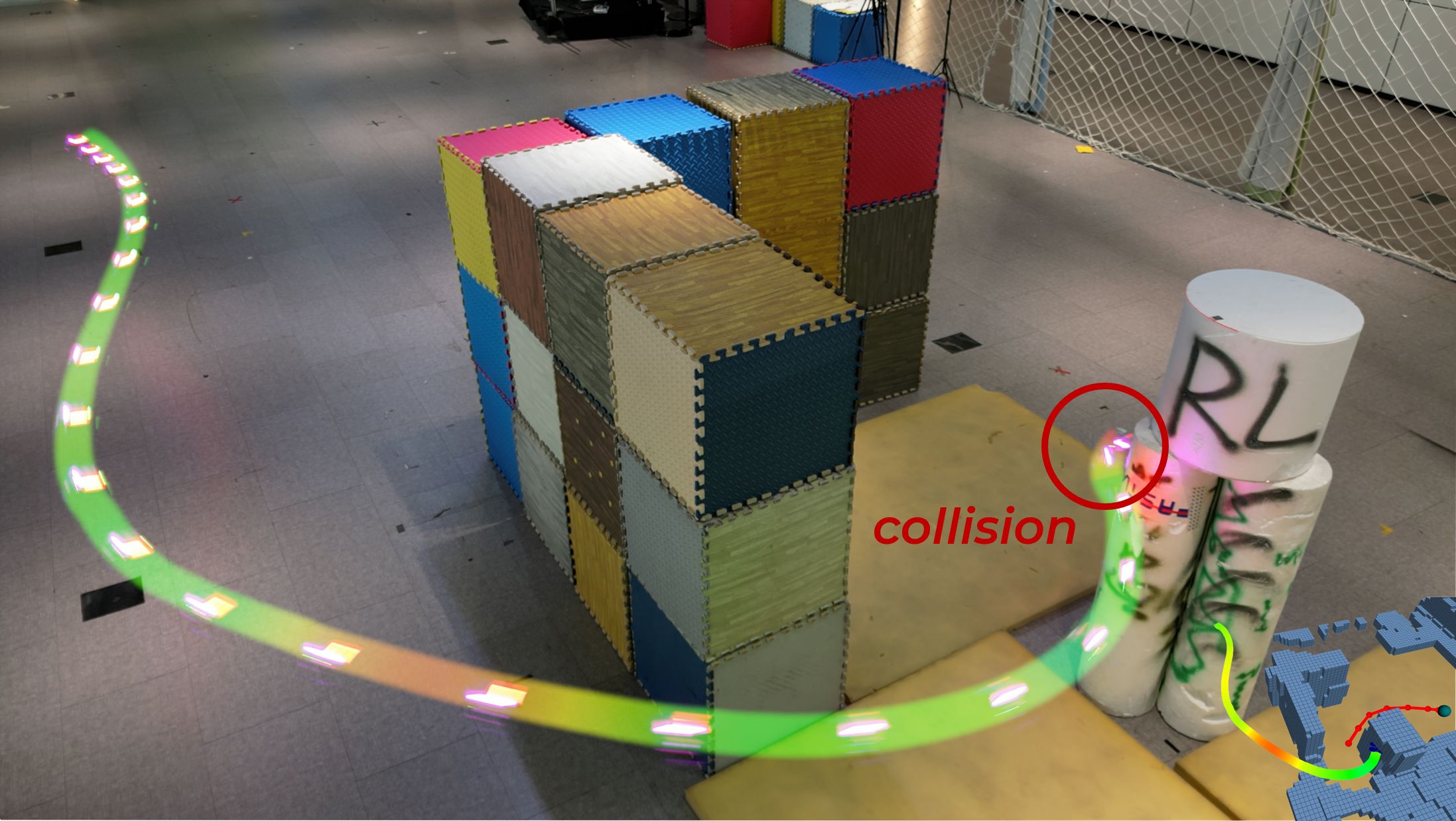}
	}      
	\hspace{-0.1cm}\subfloat[] {    
		\label{fig:3.5}
		\includegraphics[width=0.55\columnwidth]{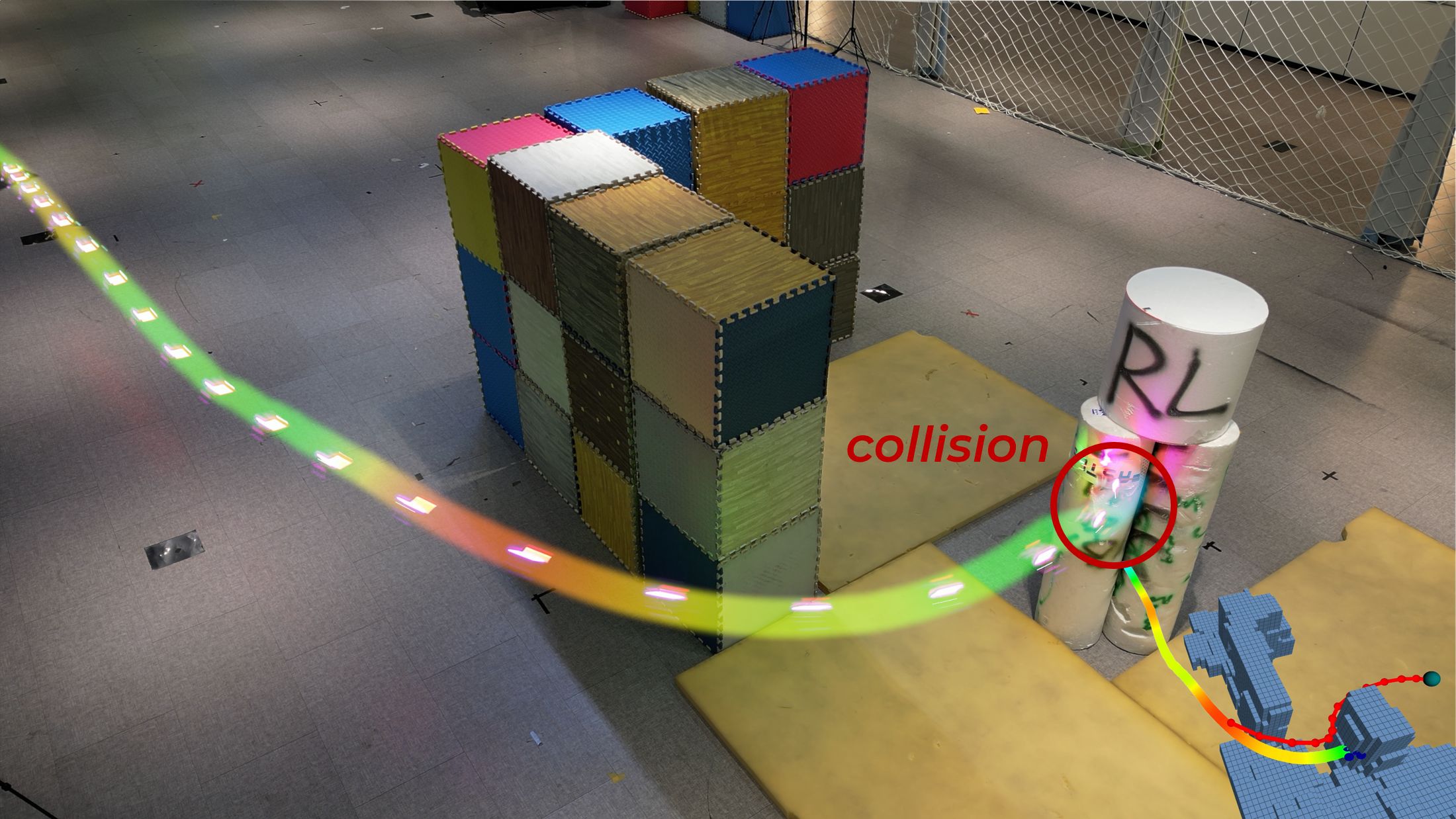}     
	} 
	\hspace{-0.1cm}\subfloat[] {    
		\label{fig:adaptation}
		\includegraphics[width=0.55\columnwidth]{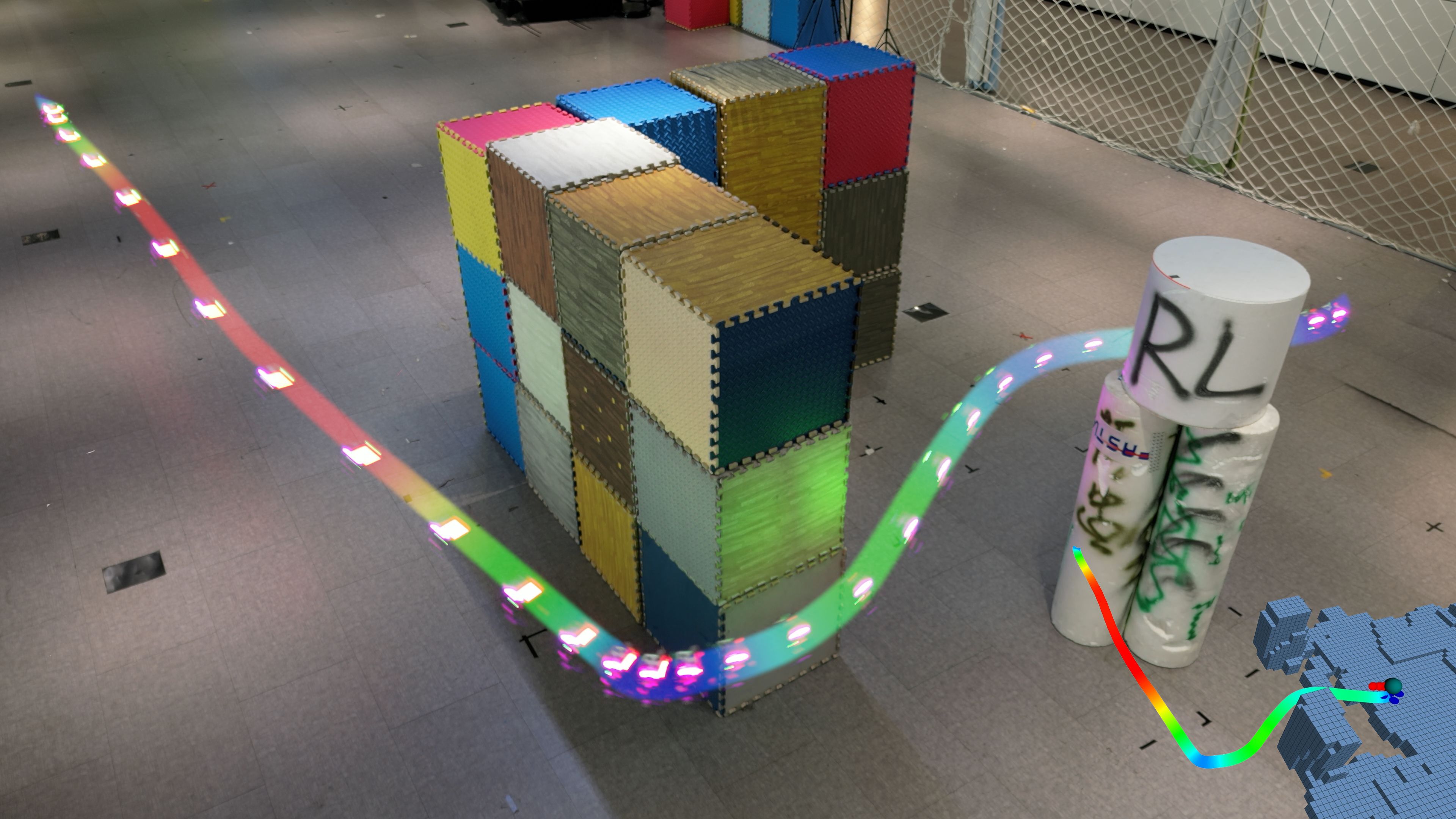}     
	}
\\  
	\captionsetup{font=footnotesize}
	\caption{ \textbf{Flight to avoid a wall-like obstacle with a hidden pillar.} The planned trajectory (red curve), the executed trajectory (colorful curve), and the local map (blue) are indicated in the bottom right corner of each figure.  (a) Flight with constant speed constraint ${v}^{\dagger}=2.5$m/s. (b)  Flight with constant speed constraint ${v}^{\dagger}=3.5$m/s (c)  Flight with speed adaptation. \vspace{-0.2cm}}

	\label{fig:wall}
\end{figure*}

\begin{figure} \centering
	\hspace{-0.15cm}
	\subfloat {  
		\raisebox{0.5cm}{\includegraphics[width=0.1\columnwidth]{legend.png}}
	}  
	\setcounter{subfigure}{0} 
	\hspace{-0.1cm}\subfloat[] {
		\label{fig:2.5-2}
		\includegraphics[width=0.3\columnwidth]{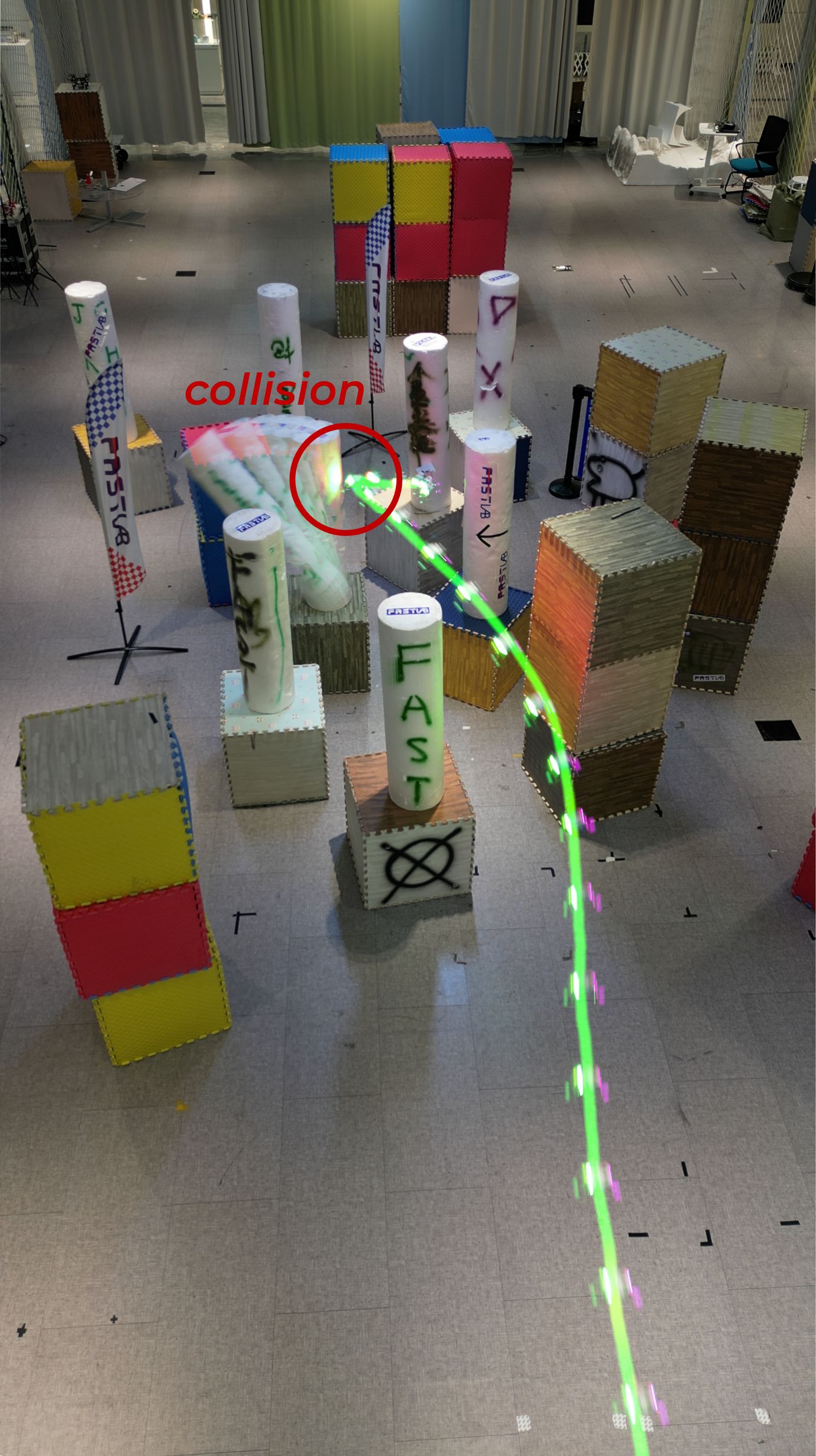}
	}      
	\hspace{-0.1cm}\subfloat[] {    
		\label{fig:adaptation-2}
		\includegraphics[width=0.3015\columnwidth]{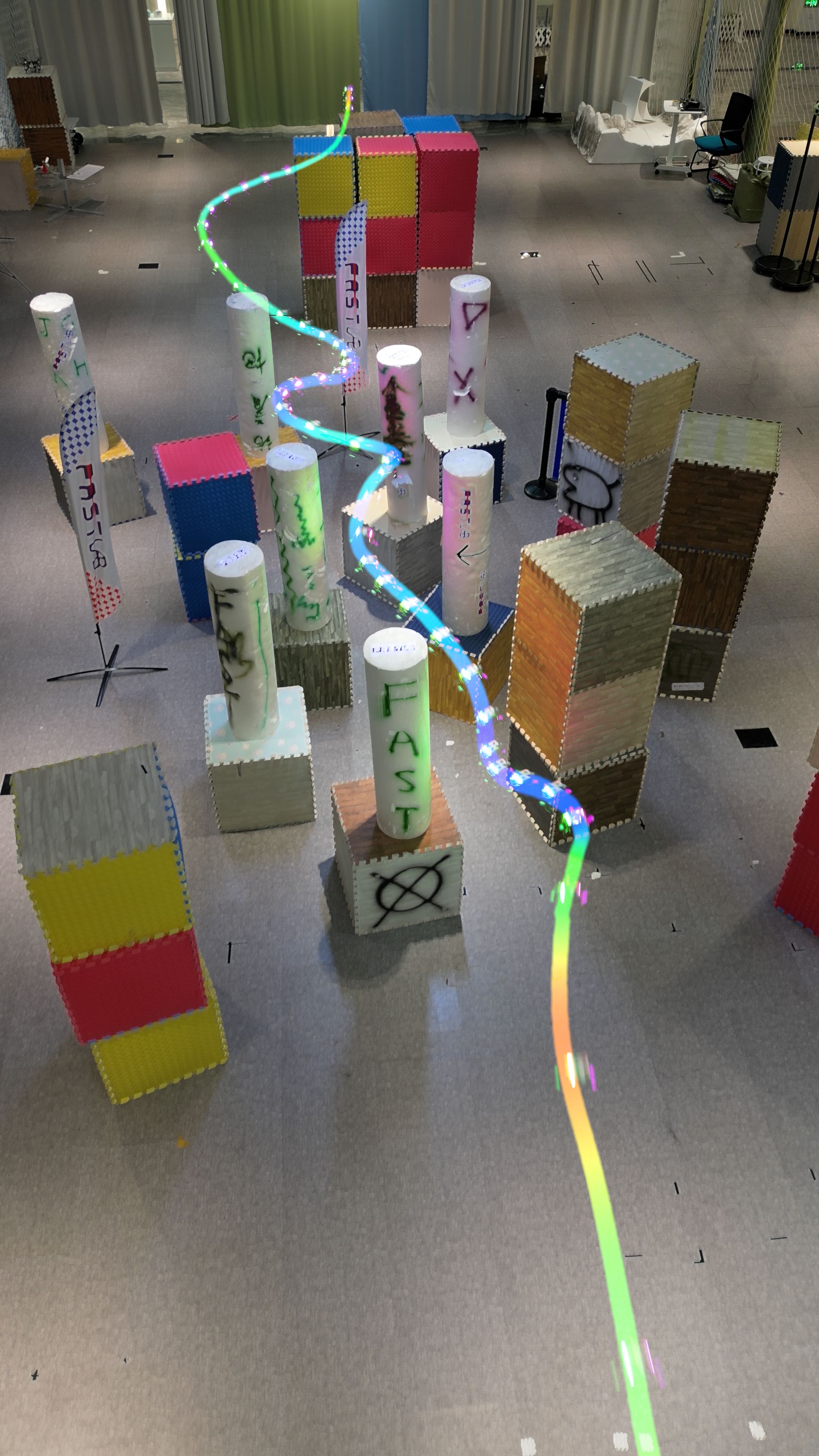}  
	}\\  
	\captionsetup{font=footnotesize}
	\caption{ \textbf{Flight through dense and large-sized obstacles.}  (a) Flight with constant speed constraint ${v}^{\dagger}=2.5$m/s. (b) Flight with speed adaptation.}
	\vspace{-0.3cm}
	\label{fig:dense}
\end{figure}
\vspace{-0.4cm}
\subsection{Real-World Experiments}

We deploy the policy on a micro drone with a size of $17$cm$\times17$cm$\times10$cm. The drone is equipped with an onboard RealSense D430 depth camera. Computation is performed on an onboard Jetson Orin NX module. We use the NOKOV motion capture system and a VIO to obtain the state estimates in indoor and outdoor scenarios, respectively. The parameter setup is aligned with that in section \ref{sec:5A}.

\subsubsection{Wall-like obstacles with an obstacle hidden behind}
We first test the policy in a representative scenario where a wall-like obstacle blocks near half of the vehicle's FOV, and only after the vehicle avoids the corner could it observe the obstacle immediately afterward.

We find that, in such a scenario, the policies under constant speed constraints, e.g.,  $v ^{\dagger}=2.5$m/s and $v ^{\dagger}_t=3.5$m/s in Fig. \ref{fig:wall}, can plan a collision-free trajectory in most of the trials. However, the mapping and planning pipeline may exhibit a significant perception latency and the planner may generate a trajectory that is not dynamically feasible. As a result, the planned trajectory, which requires sudden changes of motion, can be far from being able to be perfectly tracked by the low-level controller. We note that, typically, the trajectory is dynamically feasible and thus can be well-tracked, while in some emergency cases, the trajectory optimization method of EGO-planner \cite[Eq. (8)]{zhou2020ego}, which uses penalties to surrogate constraints for computational efficiency, can sacrifice  dynamical feasibility to ensure a collision-free trajectory.

In contrast, as shown in Fig. \ref{fig:adaptation}, after plugging the learned policy, the vehicle exhibits perception-aware behaviors, slowing down at the corner of the wall and recovering aggressiveness as more extents of the hidden obstacle is observed, until it eventually slows down to stop at the goal.  

\subsubsection{Artificial Clutters}
We evaluate the policy in an artificial scenario shown in Fig. \ref{fig:dense}. In this scenario, a dense clutter and a large-sized obstacle exist in the first and second half of the scene, respectively. The constant speed constraint policies, as shown in the Fig. \ref{fig:2.5-2}, fails in half of the trials due to planning time overruns, the occurrence of emergence stops (as explained in section \ref{sec:5C}), and also collisions caused by the same reason as that described in the previous subsection. With the learned module, the policy reasonably exhibits aggressiveness in open areas, and also behaviors cautiously among dense obstacles, as shown in Fig. \ref{fig:adaptation-2}, thus ensuring success in almost all trials. 

\subsubsection{Natural Clutters}
We also evaluate the policy in natural clutters as snapshotted in Fig. \ref{fig:head}, where the highly uneven environment triggers rich patterns of vehicle's behavior. Despite the particularly complex environment, our policy exhibits aggressive behavior in (relatively) open areas (Fig. \ref{fig:head}, top left and bottom right), but stays cautious to ensure safety when crossing narrow gap (Fig. \ref{fig:head}, top right) or among obstacles (Fig. \ref{fig:head}, bottom left). \vspace{-0.2cm}

\section{Conclusion}

We propose a hierarchical learning and planning framework for aggressiveness-adaptive flight in cluttered environments. On the one hand, the hierarchical framework allows the system to obtain a strong overall performance thanks to the existing powerful model-based trajectory planner. On the other hand, by doing so, we complement the 'missing jigsaw puzzle' in traditional trajectory planners, thus unlocking the potential of the system while freeing humans from hand-tuning labor.

The main limitation of the current system is that the spatial distribution of the vehicle is determined by the trajectory planner itself, and thus in some adversarial environments the sub-optimality of the design (e.g., choice of topology) in the planner backbone may largely limit the allowed aggressiveness in the environment. To achieve versatile aggressive flight in any environment, one solution is to give greater dominance to the learned policy while still taking advantage of safety guarantees and generalizability rendered by the traditional planner. However, when the search space is expanded, learning can take place less efficiently especially when embedded in a model-based planner where the depth rendering and simulating the system are both time-consuming. Moreover, the specialized reward scheme used in this work may not be generalized when considering a different auxiliary action space. Therefore, both scalable training environment and a unified method for effective learning are highly desired.

 
\bibliography{references.bib}

\end{document}